\documentclass[letterpaper]{article}
\usepackage{aaai}
\usepackage{times}
\usepackage{graphicx}
\usepackage{booktabs}
\usepackage{helvet}
\usepackage{courier}
\usepackage{amsmath}
\begin{document}
%
\title{Variation is the Key: A Variation-Based Framework for LLM-Generated Text Detection}
\author{Xuecong Li \and Xiaohong Li \and Qiang Hu \and Yao Zhang \and Junjie Wang\\
Tianjin University\\
No.135 Yaguan Road, Haihe Education Park\\
Tianjin, Tianjin 300350\\
}
\maketitle
\begin{abstract}
\begin{quote}
Detecting text generated by large language models~(LLMs) is crucial but challenging. Existing detectors depend on impractical assumptions, such as white-box settings, or solely rely on text-level features, leading to imprecise detection ability. In this paper, we propose a simple but effective and practical LLM-generated text detection method, VaryBalance. The core of VaryBalance is that, compared to LLM-generated texts, there is a greater difference between human texts and their rewritten version via LLMs. Leveraging this observation, VaryBalance quantifies this through mean standard deviation and distinguishes human texts and LLM-generated texts. Comprehensive experiments demonstrated that VaryBalance outperforms the state-of-the-art detectors, i.e., Binoculars, by up to 34.3\% in terms of AUROC, and maintains robustness against multiple generating models and languages.
\end{quote}
\end{abstract}

\noindent The strong text-generation capability of large language models~(LLMs) and ease of access motivate users to rely on LLMs for content creation. However, machine-generated text has increasingly permeated the human information space consequently. According to~\cite{liu_detectability_2024}, since the release of GPT-3 and ChatGPT, the number of academic papers on arXiv identified as LLM-generated has grown exponentially. Without explicit disclosure of text provenance, LLM-generated content are easily misused across news, academia, and social media~\cite{chen2023can,chen2024combating,gressel2024discussion}, leading to the spread of misinformation, academic misconduct, and phishing. Liu et al.~\cite{liu_detectability_2024} also found that even experts perform only marginally better than random guessing on identifying LLM-generated texts, suggesting that humans are not capable enough to discern LLM-generated texts anymore. As a result, developing reliable detectors for LLM-generated text is crucial. 

\begin{figure}
    \centering
    \includegraphics[width=\columnwidth]{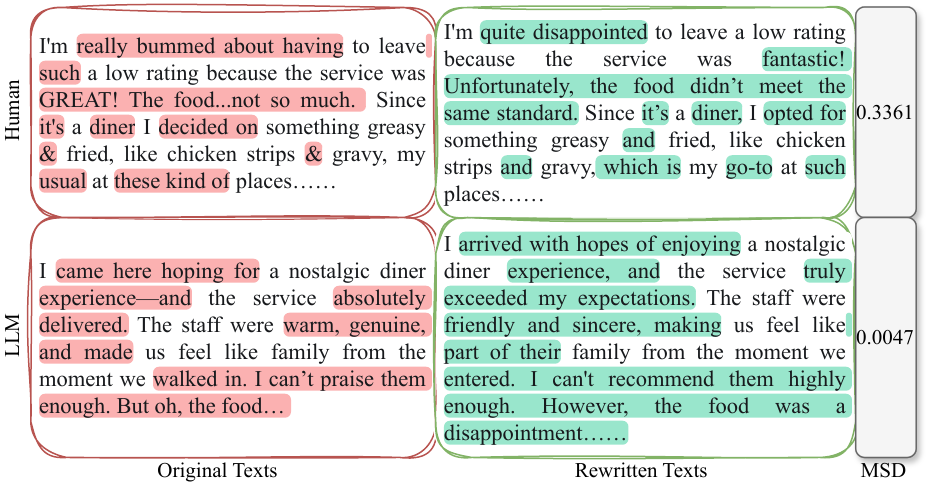}
    \caption{Comparison of Human/LLM texts and their rewritten texts.}
    \label{fig:observation_example}
\end{figure}

To address this, multiple automatic LLM-generated text detectors have been proposed, which can be roughly divided into white-box and black-box two groups. White-box methods need to know the target LLMs and access the internals of LLMs, such as logits. Representative white-box methods include DetectGPT~\cite{mitchell_detectgpt_2023} and Fast-DetectGPT~\cite{bao2023fast}. Black-box approaches make predictions based solely on observable features of the text, or information obtained from surrogate models~(not the target model). N-gram statistics~\cite{yang_dna-gpt_2023}, Levenshtein score~\cite{mao_raidar_2024}, grammar errors~\cite{wu-etal-2025-wrote}, and Neural-based methods~\cite{hu2023radar,wu2024detectrl,chen2025ipad} are commonly used black-box methods.

However, both methodologies are constrained by their inherent limitation. For white-box detectors, it is generally impractical to precisely identify the underlying text generator in real-world scenarios, and access to the generator’s logits is typically unavailable or infeasible. 
In contrast, as LLMs continue to advance, producing text that highly resembles human writing is no longer challenging, rendering these black-box methods~(such as simple text-level methods) fail to detect LLM-generated texts. 

To address these limitations, we introduce a simple but effective LLM-generated text detection method, VaryBalance. The core idea of VaryBalance is that, compared to human texts, there is a lighter difference between LLM-generated texts and their rewritten versions LLM. Figure~\ref{fig:observation_example} shows texts from humans and LLMs with their corresponding rewritten versions by GPT-4o-mini. We can see that the LLM-generated text and its rewritten version follow similar patterns, and thus are more similar after quantified by Mean Standard Deviation(MSD). Based on this observation, VaryBalance employs a rewriter to generate multiple rewrites of an input text, and a scorer to compute the log perplexity of both the original and rewritten texts. After that, it uses the log perplexity and the MSD of the rewritten texts’ log perplexities,  computed with respect to the original text’s log perplexity as the reference, as an indicator for LLM-generated content detection. Moreover, we present an extended scoring variant tailored to shorts and more stylistically diverse social media text. 

To evaluate the effectiveness of VaryBalance, we conduct comprehensive experiments on eight datasets and five models. Results demonstrate that in formal writing contexts, VaryBalance outperforms state-of-the-art methods, such as Binocular and Fast-DetectGPT, with an overall improvement of 34.5\% in terms of AUROC. 

In summary, our contributions are:

\begin{itemize}
    \item We empirically reveal that, compared to LLM-generated texts, there is a greater difference between human texts and their rewritten version via LLMs. 
    
    \item Based on our observation, we propose a novel black-box LLM-generated text detection method, VaryBalance.
    
    \item We conduct comprehensive experiments to showcase the SOTA performance of VaryBalance and its robustness under different models and languages.
\end{itemize}

\section{Related Work}

Detecting LLM-generated texts has become an increasingly challenging task. LLMs achieve superior performance on a wide range of language benchmarks and can produce convincing, contextually appropriate text~\cite{mitchell_detectgpt_2023}. Moreover. Liu et al. \cite{liu_detectability_2024} show that even domain experts perform only slightly better than random guessing when attempting to identify LLM-generated text. Hence, developing reliable detectors for LLM-generated content is both important and urgent.

We introduce approaches of white-box settings or black-box settings. There is also research~\cite{kirchenbauer2023reliability} on embedding watermarks into LLM or the generated texts for further detection. Here, we do not discuss watermark-based methods as they demand cooperation from commercial or open-source LLM providers based on their inherent requirement for accessing model generation processes or deployment infrastructure, which is often infeasible.

\subsection{White-box Methods}

What characterizes white-box approaches is the direct access to the logits of the generative model, which evaluates the LLM's confidence over every possible next token. \cite{hans2024spotting} proposed Binocular, which measures the ratio between perplexity and cross-perplexity as a detection criterion. \cite{wu2023llmdet} introduced LLMDet, a multi-model text attribution tool that constructs an $n$-gram–to–next-token probability mapping for each language model and computes proxy perplexities to classify the source of the text.

Beyond direct logit-based scoring, other white-box works leverage perturbation ~\cite{mitchell_detectgpt_2023,bao2023fast,bao_glimpse_2025}. These methods perturb text using a language model and measure changes in log-probability or other scores under the target model. 

Additional works that amend the existing works include TOCSIN~\cite{ma_zero-shot_2024}, which combines token cohesiveness with conventional logit-based detection through a dual-channel architecture, and AdaDetectGPT~\cite{zhou2025adadetectgpt}, which learns an adaptive witness function to improve log probabilities. Text Fluoroscopy~\cite{yu_text_2024} explores another methodology that exploits internal hidden representations of LMs for classification.  

White-box detectors are typically zero-shot and easy to deploy. However, they rely on a less impractical scenario in which detectors have direct access to model logits. Most commercial LLM APIs do not expose logits, and many users lack the computational resources to deploy open-source models locally.

\subsection{Black-box Methods}

Black-box detectors do not require access to the model or any proxy logits. Instead, they rely on observable textual signals. The way these methods process these features includes text truncation and continuation~\cite{yang_dna-gpt_2023}, grammar error correction~\cite{wu-etal-2025-wrote} and regeneration~\cite{mao_raidar_2024}.

Other works employ neural classifiers or fine-tuned language models. \cite{fagni_tweepfake_2021} demonstrated that fine-tuned RoBERTa achieves state-of-the-art classification performance. \cite{hu2023radar} proposed RADAR, an adversarial learning framework that jointly trains a paraphraser and a detector to achieve robust AI-text detection.\cite{liu_detectability_2024} constructed the GPABench dataset to study the detection of ChatGPT-generated academic content and trained an LSTM-based classifier. \cite{nguyen2024simllm} extended Raidar by incorporating multiple rewrites and ranking, followed by fine-tuned RoBERTa classification. ~\cite{chen2025ipad} formulates AI-generated text detection as an inverse prompt reasoning problem, and detects LLM-generated text by inverting the generation prompt and validating its consistency with the input text.

Despite diverse efforts, both white-box and black-box methods face intrinsic limitations. As LLMs continue to evolve, it is close to the world where LLM-generated texts are indistinguishable from human-written text, risking many fundamental theories of existing black-box approaches and highlighting the need for new detection paradigms.

\section{VaryBalance}


\begin{figure*}[t]
  \centering
  \includegraphics[width=.8\textwidth]{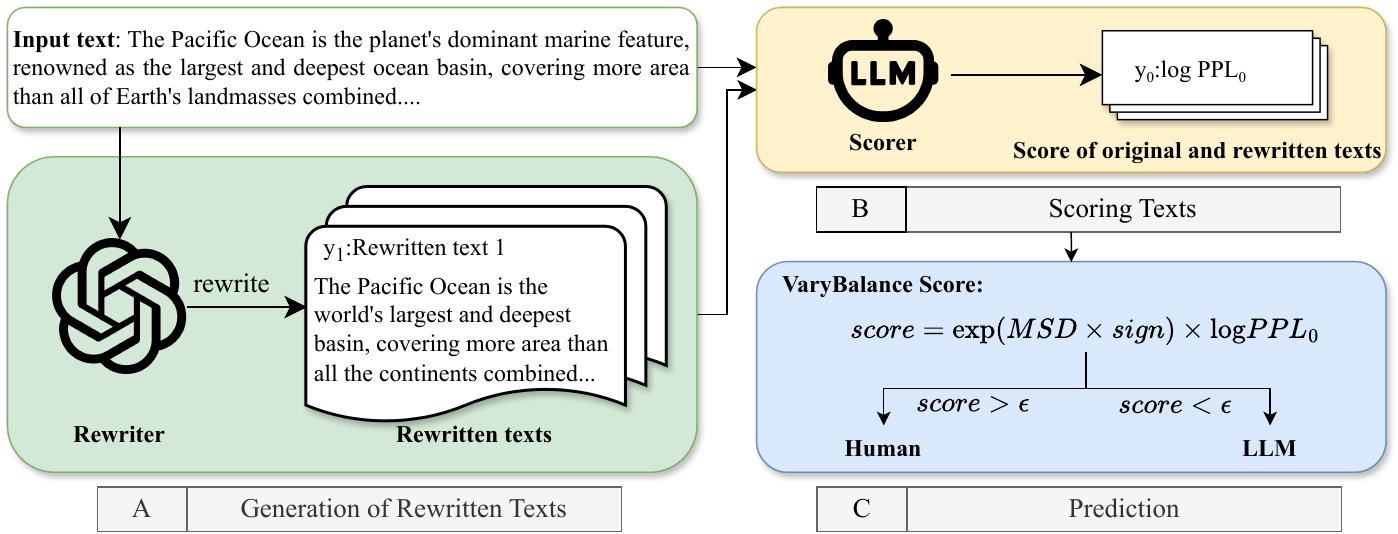}
  \caption{Overall workflow of VaryBalance. We first prompt an LLM to rewrite the text, then utilize a scoring model to calculate $\mathrm{log PPL}$ and the VaryBalance score of each text piece. The greater VaryBalance score indicates the more possibility of human text.}
  \label{fig:framework}
\end{figure*}

We propose \textbf{VaryBalance}, a black-box LLM-generated text detector. As shown in Figure~\ref{fig:framework}, VaryBalance consists of three key components: generation of rewritten texts, scoring texts, and prediction.

\paragraph{Generation of rewritten texts.} We leverage the advanced linguistic capabilities of a modern LLM to generate rewritten variants of the input text. We name the LLM Rewriter. To minimize the effect of the prompt, we only instruct the rewriter with a simple prompt: \textit{Revise this text.}

\paragraph{Scoring Texts.} After the preparation of texts and their rewritten versions, a scorer is needed to transfer these texts to computable values for further comparison. VaryBalance employs Perplexity~(PPL) as the quantification score. PPL measures how \textit{confused} a model is in predicting the next token: lower perplexity indicates higher confidence and better predictive performance, making it a useful indicator for model-generated content, and has been widely used in quantifying texts~\cite{solaiman2019release,mitchell_detectgpt_2023,bao2023fast,yang_dna-gpt_2023,hans2024spotting}

Given a token sequence $W = (w_1, w_2, \dots, w_n)$, the perplexity of a model $M$ is defined as:
\begin{equation}
\label{eq:logppl}
\mathrm{PPL}(W) = \exp \left(-\frac{1}{n} \sum_{i=1}^{n} \log P_M(w_i \mid w_{<i}) \right),
\end{equation}
where $P_M(w_i \mid w_{<i})$ denotes the conditional probability of token $w_i$ given its preceding context $w_{<i}$, 
and $n$ is the length of the token sequence. However, direct use of $\mathrm{PPL}$ is not ideal for a scoring-based detector for the instability introduced by the exponential operation. Instead, we adopt the logarithmic form $\log(\mathrm{PPL})$ as a more stable and interpretable component in our framework. VaryBalance employs a small LLM to compute the $\log(\mathrm{PPL})$ of the original texts and rewritten variants, according to Eq.~(\ref{eq:logppl}). 

\paragraph{Prediction.} The next step is to quantify the difference between texts and their rewritten variants. VaryBalance integrates Mean Standard Deviation~(MSD) for this quantification. Here, $\log(\mathrm{PPL})$ is used as the indicator, and the log perplexity of the original text $\log(\mathrm{PPL}_0)$ is selected as the central reference.

\begin{equation}
    \mathrm{MSD}(\log(\mathrm{PPL}_0)) = \frac{1}{n} \sum_{i=1}^{n} (\log(\mathrm{PPL}_i) - \log(\mathrm{PPL}_0))^2
\end{equation}

where $\log(\mathrm{PPL}_i)$ represents the log perplexities of rewritten texts, and $n$ is the total number of rewritten texts.


We found that some machine-generated texts exhibit lower log perplexity scores than the rewritten texts, as they align more closely with the preferences of the relatively smaller scoring model. This results in inflated MSD values for some of LLM-generated texts. To address this issue, we introduce Eq.~(\ref{eq:sign}) to adjust the influence of MSD on the total score. With a negative sign, a larger MSD corresponds to a stronger indication of the LLM-generated text.

\begin{equation}
    \label{eq:sign}
    \mathrm{Sign}=\mathrm{Sign}(\log(\mathrm{PPL}_0)-\frac1n\sum_{i=1}^n\log(\mathrm{PPL}_i))
\end{equation}

Finally, VaryBalance calculates the final score as:

\begin{equation}
    \label{eq:varybalance}
    \mathrm{Score}=\exp(sign \cdot \operatorname{MSD}(\log(\mathrm{PPL}_0))\cdot \log(\mathrm{PPL}_0)
\end{equation}

We describe the influence of MSD on the original text’s log perplexity using an exponential function, which allows human-written text to fully benefit from the amplification effect of MSD, thereby creating a clear separation from machine-generated text with similar log perplexity values. The score is then compared with a chosen threshold to classify LLM-generated text.

We also offer an expansion of the original VaryBalance score. We observe that for social media–style text, the gap in MSD between human-written text and the variance of its rewritten texts is substantially larger than that of machine-generated text. This discrepancy can be leveraged to refine the original $\mathrm{Score}$ for social media texts for boosting the detection performance. 

\begin{equation}
    \label{eq:extend_varybalance}
    \begin{aligned}
    \mathrm{Score_e}&=\exp(sign \cdot \rho\cdot\mathrm{MSD}(\log(\mathrm{PPL}_0)))\cdot \log(\mathrm{PPL}_0)\\
    \rho&=\frac{\mathrm{MSD}(\log(\mathrm{PPL}_0)}{\mathrm{Var}({\log(\mathrm{PPL}_i)) \wedge 1 \leq  i \leq n})}
    \end{aligned}
\end{equation}

where $\mathrm{Var}$ is the variance, and $\rho$ as the expansion coefficient that is applied to texts with a more informal style.

\section{Preliminary Study}

\begin{figure}
    \centering
    \includegraphics[width=\linewidth]{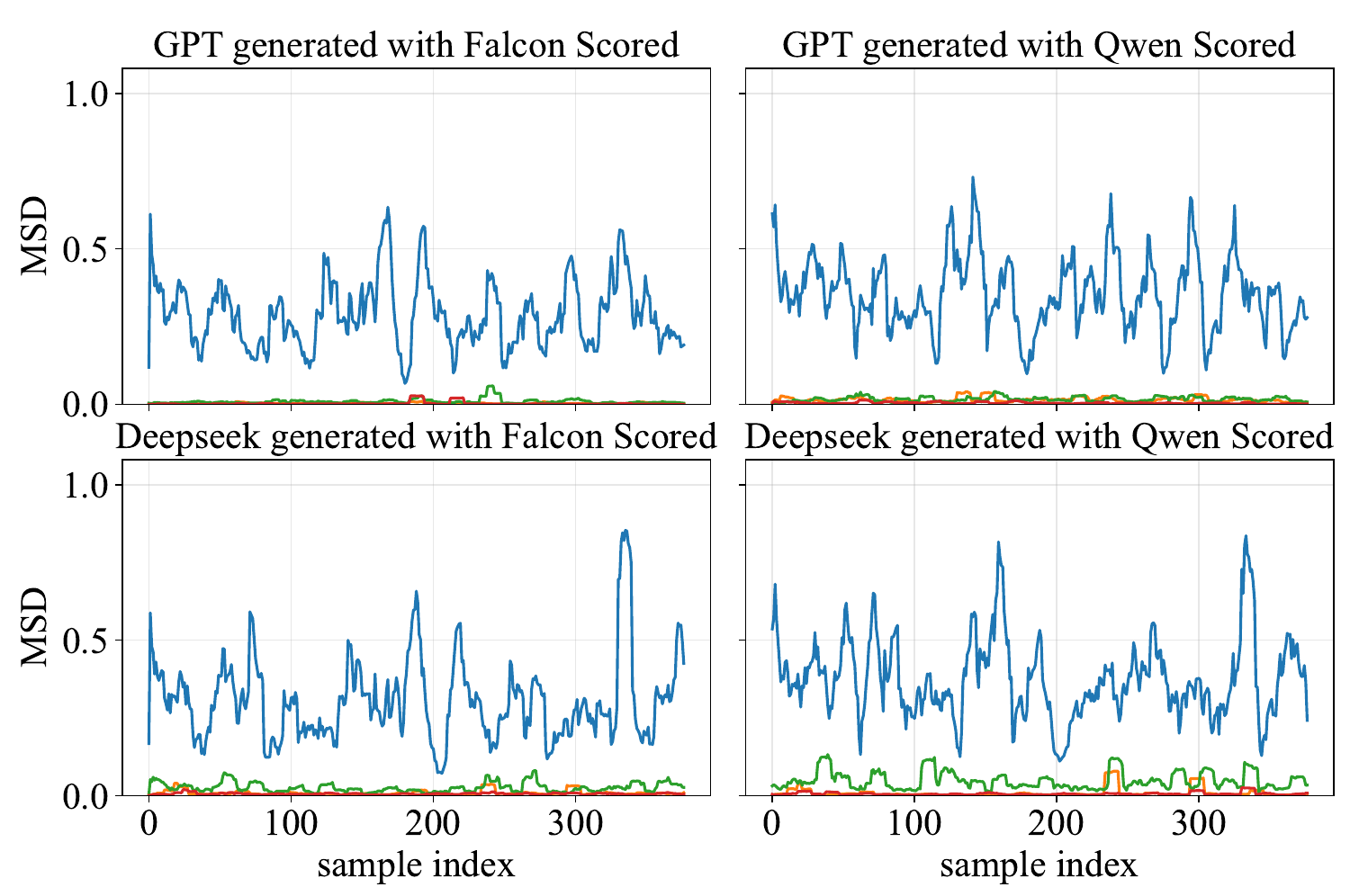}
    \caption{The MSD of human text, machine text, human text rewritten by LLM ,and machine text rewritten by LLM, Human text indicates texts are written by humans, the logic suits for machine text etc. The blue line represents human text.}
    \label{fig:observation}
\end{figure}

The proposed new method is motivated by the observation that, compared to LLM-generated texts, there is a greater difference between human texts and their rewritten version via LLMs. To systematically validate this assumption, we conduct a preliminary study.

Specifically, we first randomly sample 400 text pairs from the HC3 dataset \cite{guo-etal-2023-hc3}. Each pair in the dataset consists of a question with both a human-written answer and a machine-generated answer. Then, we use these questions as prompts to produce new machine-generated responses, and rewrite machine-generated responses and original human texts using LLMs. As a result, we have two groups of data, human-generated texts with their rewritten versions, and machine-generated texts with their rewritten versions. After that, we employ MSD to quantify the difference between texts and their rewritten versions.

We consider two types of LLMs for the rewritten, DeepSeek-chat and GPT-4o-mini, and two types of LLMs as scorers, Falcon and Qwen. Fig.~\ref{fig:observation} depicts the results. We can see that it is clear that the MSD of human text and its rewritten texts is greater than that of the counterpart of LLM-generated texts and their rewritten texts. The average MSD value of human texts is around 0.34, while that of machine texts is around 0.009. Moreover, we observe that 96\% of total texts in the dataset follow this observation, which demonstrates that our assumption stands. 


\begin{figure*}[t]
  \centering
  \includegraphics[width=\textwidth]{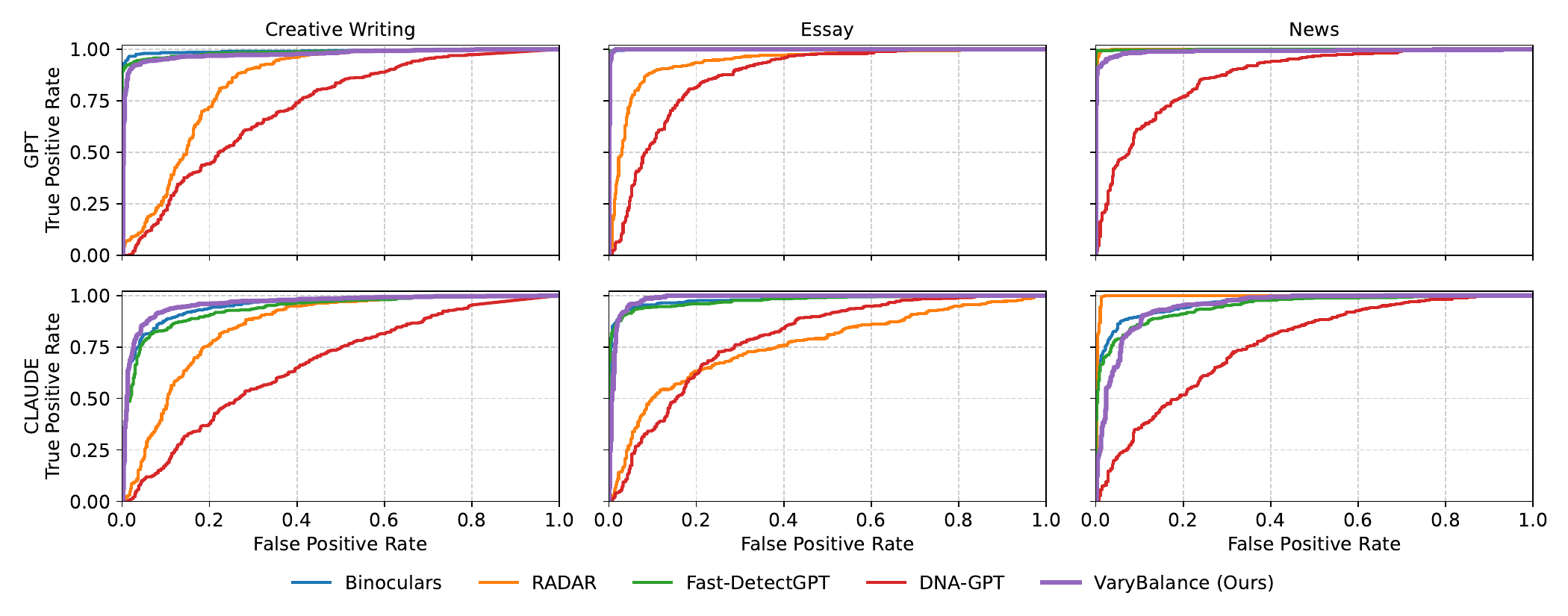}
  \caption{The ROC curve of VaryBalance and the other four baselines on the benchmark datasets. Upper: GPT as the source model; Lower: Claude as the source model.}
  \label{fig:benchmark}
\end{figure*}

To assess the effectiveness of VaryBalance, we conduct comparative experiments across multiple languages, generation models, generation strategies, and application scenarios. It is worth noting that the part of the dataset used to design the VaryBalance scoring function is not involved in the performance evaluation of VaryBalance.

\subsection{Datasets}

\paragraph{Benchmarks.}We evaluate our method on several datasets that are widely used in prior literature. Among them, a commonly adopted benchmark introduced by \cite{verma_ghostbuster_2024} consists of three datasets: Creative Writing, News, and Student Essay. For each dataset, the authors generate machine-written counterparts of the human texts using gpt-3.5-turbo as the training set, while additional texts generated by Claude are provided as a held-out test set.


\paragraph{Datasets for Robustness Experiment} We select three datasets to represent different genres of texts with different lengths. Drawing samples from dataset CNN~\cite{hermann2015teaching}, reddit posts dataset RedditClean~\footnote{https://huggingface.co/datasets/SophieTr/reddit\_clean} and yelp review dataset YelpReviewFull~\cite{zhang2015character}, we generate corresponding LLM-generated texts using Qwen3-Next-80B-A3B-Instruct~\cite{qwen2.5-1m,qwen3technicalreport} and grok-4-fast~\footnote{https://x.ai/news/grok-4-fast}. All human texts were collected before the release of ChatGPT, which ensures that they are not contaminated by LLM-generated content.

\paragraph{Multi-language.}We evaluate the multilingual capability of our detector using the M4 dataset\cite{wang-etal-2024-m4}, which contains texts in Chinese, Arabic, Urdu, and Bulgarian collected from multiple sources. We regenerate the machine-generated texts with qwen3-235b-a22b-instruct-2507\cite{qwen2.5-1m,qwen3technicalreport} by following the prompts provided in the dataset, thereby better simulating outputs from more recent and capable large language models. To reduce bias towards non-native speaker, we additionally employ the EssayForum\footnote{https://huggingface.co/datasets/nid989/EssayFroum-Dataset} dataset, which contains both original student-written essays and their grammar-corrected counterparts.

\paragraph{Rewrite.}To evaluate the detector’s ability to identify rewritten text, we construct a rewritten-text dataset based on the HC3 dataset\cite{guo-etal-2023-hc3} using two large language models, deepseek-chat and qwen3-235b-a22b-instruct-2507. 

\subsection{Baseline}

Among the white-box detectors, we compare our detector with Fast-DetectGPT\cite{bao2023fast}, Binoculars\cite{hans2024spotting} and AdaDetectGPT~\cite{zhou2025adadetectgpt}. For black-box detectors, we choose RADAR\cite{hu2023radar} and DNA-GPT\cite{yang_dna-gpt_2023} as baselines. RADAR presents a neural-based detector trained through adversarial learning for robust AI-text detection, while DNA-GPT is mainly based on word-level features caused by text continuation through an LLM.

\subsection{Metric}

Following previous works~\cite{mitchell_detectgpt_2023,ma_zero-shot_2024,zhou2025adadetectgpt}, we evaluate the detection effectiveness using the area under the receiver operating characteristic curve (AUROC).AUROC indicates the probability of a detector to rank a randomly-selected positive sample higher than a negative counterpart, capturing the overall performance of the subject.
\section{Experiment}

\subsection{Results}
\begin{figure}
    \centering
    \includegraphics[width=\columnwidth]{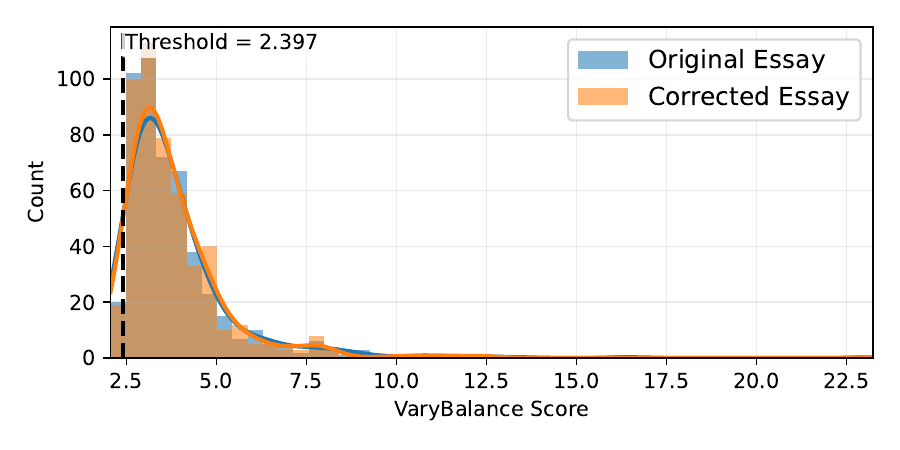}
    \caption{The comparison of score distribution of VaryBalance on Essay Forum dataset.}
    \label{fig:nonnative}
\end{figure}
\begin{figure*}[t]
  \centering
  \includegraphics[width=\textwidth]{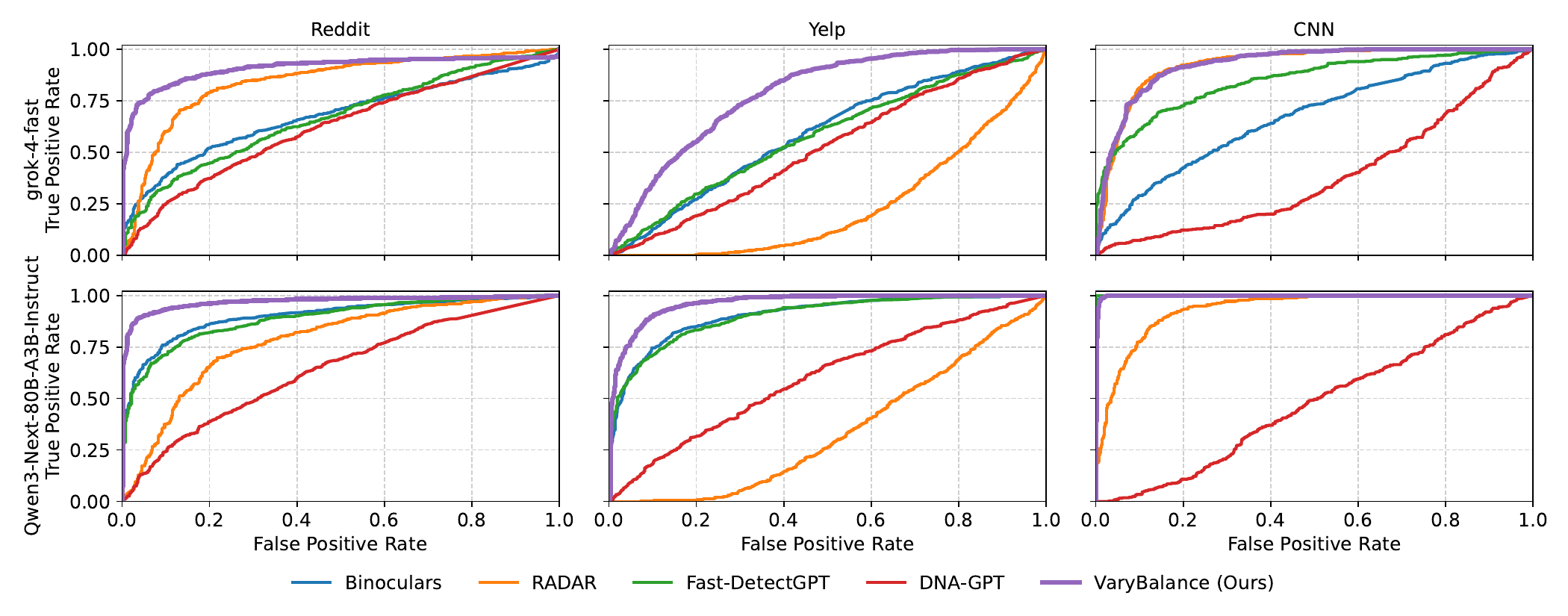}
  \caption{The ROC curve of VaryBalance and the other five baselines on the CNN, Reddit and Yelp. Upper: grok as the source model; Lower: qwen as the source model.}
  \label{fig:otherroc}
\end{figure*}
\begin{table}[t]
    \centering
    \small
    \begin{tabular}{lrrr}
        \toprule Detectors  & Grok         & Qwen      & GPT \\
        \midrule
        DNA-GPT                    & 0.6216      & 0.6406     & 0.7035  \\
        RADAR                      & 0.8413          & 0.7817      & 0.9842  \\
        Binoculars                 & 0.6763          & 0.9009     & 0.9846 \\
        Fast-DetectGPT             & 0.6683          & 0.8876     & 0.9623   \\
        AdaDetectGPT                &0.2693         &0.6352         &0.3285\\
        VaryBalance                & 0.9083         & 0.9491      & 0.9870 \\
        VaryBalance with expansion & \textbf{0.9342} & \textbf{0.9862} & \textbf{0.9936} \\
        \bottomrule
    \end{tabular}
    \caption{AUROC score of different detectors on Reddit under Grok, Qwen and GPT generations.}
    \label{tab:result on reddit}
\end{table}
\subsubsection{Benchmark Performance}
As shown in Fig.~\ref{fig:benchmark}, VaryBalance consistently outperforms DNA-GPT, RADAR, and AdaDetectGPT across most experimental settings, while achieving performance comparable to Fast-DetectGPT and Binoculars. When the generation model shifts to Claude, nearly all detectors exhibit a slight performance degradation; nevertheless, VaryBalance maintains a high level of effectiveness and continues to surpass the baseline methods.

It is worth noting that the machine-generated portions of the Creative Writing, Essay, and News datasets are produced using gpt-3.5-turbo, which is relatively less capable and slower compared to more recent models. Evaluating performance solely on these datasets is insufficient.

\subsubsection{Performance on Various Text Sources and Genres}

In real-world applications, LLM-generated text exhibits substantial variation in length, style, domain, and generating models. Therefore, we evaluate our detector on the Reddit, Yelp, and CNN datasets, which represent different genres of texts with different lengths from different source models, enabling us to assess the detector’s robustness to LLM-generated content beyond the GPT and Claude model families.
The result shown in Fig.~\ref{fig:otherroc} demonstrates that VaryBalance surpasses baselines by a substantial performance gap on Yelp and Reddit datasets while maintaining comparable effectiveness with baselines on CNN datasets. Combining the benchmark results, we observe that the baselines perform well on relatively formal and longer texts (e.g., mainstream news) but exhibit pronounced performance degradation on social media-style text. Our detector remains robust to the change in text genres with an AUROC of 0.949 on Reddit, 0.963 on Yelp, and 0.997 on CNN, surpassing other detectors by margins of 0.30.

In terms of the impact of generating models, we perform comparative evaluations on Reddit texts generated by three models, Grok, Qwen, and gpt-4o-mini(represented as GPT), with the results summarized in Table~\ref{tab:result on reddit}. VaryBalance already outperforms baselines, and the VaryBalance score with expansion further enhances the advantages. We therefore conclude that VaryBalance is robust to generating models and text genres.
\begin{table*}
    \centering
    \small
    \begin{tabular}{lrrrrrrrr}
        \toprule
        & \multicolumn{4}{c}{ChatGPT} & \multicolumn{4}{c}{Qwen3-235b-a22b-instruct} \\
        \cmidrule(lr){2-5} \cmidrule(lr){6-9}
        Detectors 
        & Arabic & Bulgarian & Urdu & Chinese
        & Arabic & Bulgarian & Urdu & Chinese \\
        \midrule
        DNA-GPT        &0.2117  &0.3145  &0.4571  &0.6310  &0.6780  &0.5904  &0.4571  &0.6310  \\
        RADAR          &0.3570 &0.6969  &0.4462  &0.3631  &0.4725  &0.7842  &0.4462  &0.3631  \\
        binoculars     &0.8355  &0.8658  &0.9420  &0.7749  &0.7680  &0.6280  &0.9421  &0.7749  \\
        fast-detectgpt &0.6146  &0.8692  &0.7846  &0.4416  &0.6170  &0.4466  &0.7847  &0.4416  \\
        AdaDetectGPT   &0.5632&\textbf{0.8847}   &0.6784  &0.4728  &0.9489  &0.8851  &0.6527  &0.9148
        \\
        VaryBalance    &\textbf{0.8357}&0.8821  &\textbf{0.9867}  &\textbf{0.8581}  &\textbf{0.9708}  &\textbf{0.9444}  &\textbf{0.8917}  &\textbf{0.9752}  \\
    \bottomrule
    \end{tabular}
    \caption{The AUROC scores on detecting the four languages where ChatGPT and Qwen3-235b-a22b-instruct are generating models. The AUROC scores are rounded to four decimal places.}
    \label{tab:result on multi-lan}
\end{table*}
\subsubsection{Multi-language Scenarios}

Our detector nearly achieves the best performance across all evaluated languages. As shown in Table~\ref{tab:result on multi-lan}, on ChatGPT-generated texts, VaryBalance attains state-of-the-art performance, with an AUROC of up to 0.986. On text generated by Qwen, VaryBalance consistently achieves AUROC values above 0.892, with the highest reaching 0.975, significantly surpassing the performance of other methods.

An important challenge in machine-generated text detection is that some detectors exhibit a bias toward classifying texts written by non-native English speakers as machine-generated \cite{liang2023gpt}. On the EssayForum dataset, VaryBalance achieves an accuracy of 0.98. Meanwhile, Figure.~\ref{fig:nonnative} shows that the score distributions of the original essay and grammar-corrected essay are highly similar, indicating that our detector does not exhibit bias against text written by non-native English speakers.

\subsubsection{Performance on Rewritten Texts}
\begin{table}
    \centering
    \small
    \begin{tabular}{lrrr}
        \toprule Detectors  & GPT         & Qwen      & DS \\
        \midrule
        DNA-GPT                    & 0.5151      & 0.4963     & 0.5061  \\
        RADAR                      & 0.3570          & 0.3641      & 0.3642  \\
        Binoculars                 & 0.8225          & 0.6272     & 0.5882 \\
        Fast-DetectGPT             & 0.8103          & 0.6004     & 0.5671   \\
        AdaDetectGPT               &0.7673          &0.5980             &0.6288\\
        VaryBalance                & 0.9098         & 0.8919      & 0.8200 \\
        VaryBalance with expansion & \textbf{0.9299}  & \textbf{0.9249}     & \textbf{0.8438} \\
        \bottomrule
    \end{tabular}
    \caption{The AUROC on the Rewrite dataset. GPT: GPT-4o-mini, Qwen: Qwen3-Next-80B-A3B-Instruct, DS: deepseek-chat. The AUROC scores are rounded to four decimal places.}
    \label{tab:result on rewrite}
\end{table}
Prior studies have shown that machine-rewritten text is particularly difficult to detect\cite{wahle-etal-2022-large,pudasaini2024survey}. 
As shown in Table~\ref{tab:result on rewrite}, VaryBalance significantly outperforms baselines: across three different models, it consistently surpasses other detectors, achieving AUROC scores above 0.82 in all cases. Moreover, VaryBalance with expansion further increases AUROC by 0.0256 on average across three settings. 

Overall, the experimental results show that VaryBalance is robust to variations in text type, length, generation model, and language. It effectively identifies LLM-rewritten text and, through the incorporation of a single parameter, further enhances performance on shorter, social media–style texts, surpassing current state-of-the-art detection models.
\begin{figure*}
\centering
  \includegraphics[width=\textwidth]{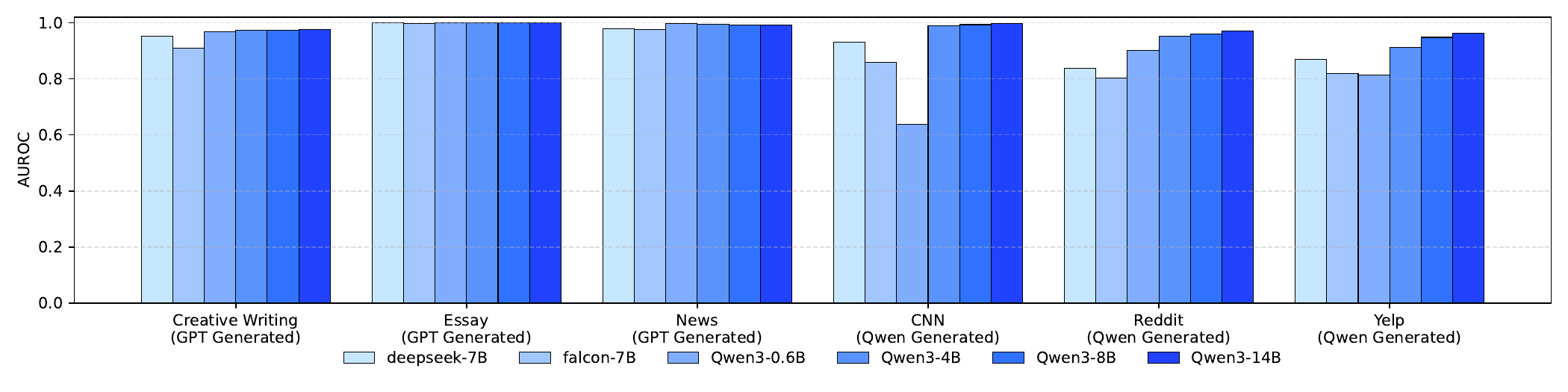}
  \caption{Comparison on performance of different series and size of scoring model.}
  \label{fig:size}
\end{figure*}

\section{Discussion}

\begin{figure}
    \centering
    \includegraphics[width=\linewidth]{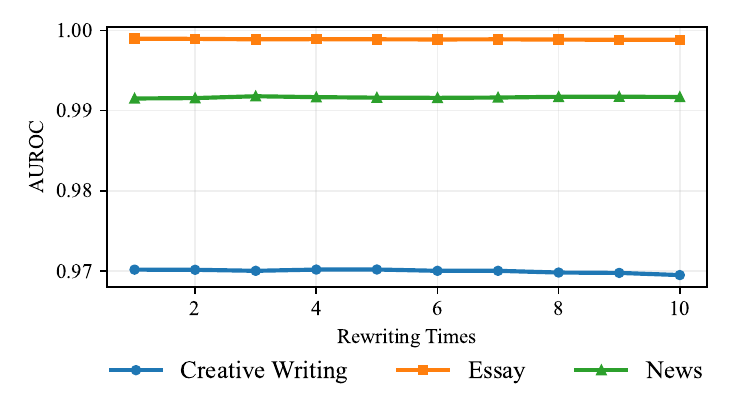}
    \caption{The AUROC scores are stable regardless of regeneration times.}
    \label{fig:rewrite_times}
\end{figure}
We investigate the contribution of each component in VaryBalance to its overall performance, including the scoring function, the number of rewrites, the text length, and the choice of the scoring model.
\begin{table}[t]
    \centering
    \small
    \begin{tabular}{lrrr}
        \toprule Scores          & Reddit         & Essay      & Rewrite \\
        \midrule
        log PPL                    & 0.9421      & \textbf{0.9989}     & 0.8119  \\
        VaryBalance                & 0.9692      & \textbf{0.9989}      & 0.9097  \\
        VaryBalance with expansion & \textbf{0.9861} & 0.9964     & \textbf{0.9299} \\
        \bottomrule
    \end{tabular}
    \caption{The AUROC score of log PPL, VaryBalance and VaryBalance with expansion.}
    \label{tab:influence in msd}
\end{table}
\subsection{Size and Series of Scorers}

We evaluate the impact of the scoring model on VaryBalance by selecting Qwen3 models with parameter sizes ranging from 0.6B to 14B, as well as several 7B models from other families, and applying the base VaryBalance framework across six datasets. As shown in the Fig.~\ref{fig:size}, on benchmarks, the choice of the scoring model has only a limited effect on performance, while a more pronounced performance gap shows on other datasets that were constructed using more recent models. On these datasets, we observe a clear trend of performance improvement with increasing model size. Among the 7B-scale models, DeepSeek-7B consistently outperforms Falcon-7B, while the slightly larger Qwen3-8B consistently surpasses both DeepSeek-7B and Falcon-7B. Notably, on the Reddit and benchmark datasets, even the smallest Qwen3-0.6B model outperforms DeepSeek-7B and Falcon-7B.

These results indicate that the choice of the scoring model has a significant impact on the performance of VaryBalance, encompassing both the selection of the model family and the model size. Due to computational constraints, we do not evaluate larger models; however, the experimental findings suggest that, when computational resources permit, larger Qwen models can further enhance the effectiveness of VaryBalance. The impact of model selection for the scorer is more pronounced when evaluated on texts generated by recent large language models.

\subsection{Regeneration Times}

The experiment shows that the number of regeneration times has only a marginal effect on overall performance, see Fig.\ref{fig:rewrite_times}. We hypothesize that multiple rewritten outputs produced by the same LLM tend to have similar log perplexity values, resulting in low variance among the rewritten texts and consequently a stable MSD. Therefore, in practical applications, the number of rewrites can be selected flexibly according to API constraints without significantly affecting the performance of VaryBalance.
\subsection{The influence of MSD}

To verify the effectiveness of our modification to log PPL, we conduct comparative experiments on the Reddit, Essay, and Rewrite datasets, which respectively represent social media text, academic text, and machine-rewritten text. As shown in Table.~\ref{tab:influence in msd}, both variants of the VaryBalance score achieve better performance than log PPL score, and VaryBalance with expansion yields noticeable improvements on the Reddit and Rewrite dataset, while exhibiting a slight performance degradation on the Essay dataset. These results confirm the effectiveness of our modification to log PPL, and indicate that the impact of MSD is less pronounced for longer and more formal texts, and therefore its contribution should not be overly amplified in such settings.

\section{Conclusion}

We proposed VaryBalance, a black-box LLM-generated text detector based on the observation that the difference between the MSD of human text and machine text is significant. Extensive experiments show that VaryBalance surpasses SOTA detectors by 0.3 in terms of the AUROC score while keeping robustness against multiple genres, multiple generating models, and multiple languages. 

\bibliographystyle{aaai} 
\bibliography{ref}

@inproceedings{liu_detectability_2024,
	address = {New York, NY, USA},
	series = {{CCS} '24},
	title = {On the {Detectability} of {ChatGPT} {Content}: {Benchmarking}, {Methodology}, and {Evaluation} through the {Lens} of {Academic} {Writing}},
	isbn = {979-8-4007-0636-3},
	shorttitle = {On the {Detectability} of {ChatGPT} {Content}},
	url = {https://dl.acm.org/doi/10.1145/3658644.3670392},
	doi = {10.1145/3658644.3670392},
	language = {en-US},
	urldate = {2024-12-16},
	booktitle = {Proceedings of the 2024 on {ACM} {SIGSAC} {Conference} on {Computer} and {Communications} {Security}},
	publisher = {Association for Computing Machinery},
	author = {Liu, Zeyan and Yao, Zijun and Li, Fengjun and Luo, Bo},
	month = dec,
	year = {2024},
	pages = {2236--2250},
}

@article{chen2023can,
  title={Can llm-generated misinformation be detected?},
  author={Chen, Canyu and Shu, Kai},
  journal={arXiv preprint arXiv:2309.13788},
  year={2023}
}

@article{fagni_tweepfake_2021,
	title = {{TweepFake}: {About} detecting deepfake tweets},
	volume = {16},
	issn = {1932-6203},
	shorttitle = {{TweepFake}},
	url = {https://journals.plos.org/plosone/article?id=10.1371/journal.pone.0251415},
	doi = {10.1371/journal.pone.0251415},
	language = {en},
	number = {5},
	urldate = {2025-01-14},
	journal = {PLOS ONE},
	author = {Fagni, Tiziano and Falchi, Fabrizio and Gambini, Margherita and Martella, Antonio and Tesconi, Maurizio},
	month = may,
	year = {2021},
	note = {Publisher: Public Library of Science},
	pages = {e0251415},
}

@inproceedings{yu_text_2024,
	address = {Miami, Florida, USA},
	title = {Text {Fluoroscopy}: {Detecting} {LLM}-{Generated} {Text} through {Intrinsic} {Features}},
	shorttitle = {Text {Fluoroscopy}},
	url = {https://aclanthology.org/2024.emnlp-main.885/},
	doi = {10.18653/v1/2024.emnlp-main.885},
	urldate = {2025-02-20},
	booktitle = {Proceedings of the 2024 {Conference} on {Empirical} {Methods} in {Natural} {Language} {Processing}},
	publisher = {Association for Computational Linguistics},
	author = {Yu, Xiao and Chen, Kejiang and Yang, Qi and Zhang, Weiming and Yu, Nenghai},
	editor = {Al-Onaizan, Yaser and Bansal, Mohit and Chen, Yun-Nung},
	month = nov,
	year = {2024},
	pages = {15838--15846},
}

@article{hans2024spotting,
  title={Spotting llms with binoculars: Zero-shot detection of machine-generated text},
  author={Hans, Abhimanyu and Schwarzschild, Avi and Cherepanova, Valeriia and Kazemi, Hamid and Saha, Aniruddha and Goldblum, Micah and Geiping, Jonas and Goldstein, Tom},
  journal={arXiv preprint arXiv:2401.12070},
  year={2024}
}

@misc{yang_dna-gpt_2023,
	title = {{DNA}-{GPT}: {Divergent} {N}-{Gram} {Analysis} for {Training}-{Free} {Detection} of {GPT}-{Generated} {Text}},
	shorttitle = {{DNA}-{GPT}},
	url = {http://arxiv.org/abs/2305.17359},
	doi = {10.48550/arXiv.2305.17359},
	language = {en-US},
	urldate = {2025-02-20},
	publisher = {arXiv},
	author = {Yang, Xianjun and Cheng, Wei and Wu, Yue and Petzold, Linda and Wang, William Yang and Chen, Haifeng},
	month = oct,
	year = {2023},
}

@inproceedings{mitchell_detectgpt_2023,
	title = {{DetectGPT}: {Zero}-{Shot} {Machine}-{Generated} {Text} {Detection} using {Probability} {Curvature}},
	shorttitle = {{DetectGPT}},
	url = {https://proceedings.mlr.press/v202/mitchell23a.html},
	language = {en},
	urldate = {2025-01-15},
	booktitle = {Proceedings of the 40th {International} {Conference} on {Machine} {Learning}},
	publisher = {PMLR},
	author = {Mitchell, Eric and Lee, Yoonho and Khazatsky, Alexander and Manning, Christopher D. and Finn, Chelsea},
	month = jul,
	year = {2023},
	note = {ISSN: 2640-3498},
	pages = {24950--24962},
}

@misc{mao_raidar_2024,
	title = {Raidar: {geneRative} {AI} {Detection} {viA} {Rewriting}},
	shorttitle = {Raidar},
	url = {http://arxiv.org/abs/2401.12970},
	doi = {10.48550/arXiv.2401.12970},
	language = {en-US},
	urldate = {2025-02-27},
	publisher = {arXiv},
	author = {Mao, Chengzhi and Vondrick, Carl and Wang, Hao and Yang, Junfeng},
	month = apr,
	year = {2024},
	note = {arXiv:2401.12970 [cs]},
}

@inproceedings{ma_zero-shot_2024,
	address = {Miami, Florida, USA},
	title = {Zero-{Shot} {Detection} of {LLM}-{Generated} {Text} using {Token} {Cohesiveness}},
	url = {https://aclanthology.org/2024.emnlp-main.971/},
	doi = {10.18653/v1/2024.emnlp-main.971},
	language = {en-US},
	urldate = {2025-05-01},
	booktitle = {Proceedings of the 2024 {Conference} on {Empirical} {Methods} in {Natural} {Language} {Processing}},
	publisher = {Association for Computational Linguistics},
	author = {Ma, Shixuan and Wang, Quan},
	editor = {Al-Onaizan, Yaser and Bansal, Mohit and Chen, Yun-Nung},
	month = nov,
	year = {2024},
	pages = {17538--17553},
}

@misc{verma_ghostbuster_2024,
	title = {Ghostbuster: {Detecting} {Text} {Ghostwritten} by {Large} {Language} {Models}},
	shorttitle = {Ghostbuster},
	url = {http://arxiv.org/abs/2305.15047},
	doi = {10.48550/arXiv.2305.15047},
	urldate = {2025-11-06},
	publisher = {arXiv},
	author = {Verma, Vivek and Fleisig, Eve and Tomlin, Nicholas and Klein, Dan},
	month = apr,
	year = {2024},
	note = {arXiv:2305.15047 [cs]},
}

@article{bao2023fast,
  title={Fast-detectgpt: Efficient zero-shot detection of machine-generated text via conditional probability curvature},
  author={Bao, Guangsheng and Zhao, Yanbin and Teng, Zhiyang and Yang, Linyi and Zhang, Yue},
  journal={arXiv preprint arXiv:2310.05130},
  year={2023}
}

@article{kirchenbauer2023reliability,
  title={On the reliability of watermarks for large language models},
  author={Kirchenbauer, John and Geiping, Jonas and Wen, Yuxin and Shu, Manli and Saifullah, Khalid and Kong, Kezhi and Fernando, Kasun and Saha, Aniruddha and Goldblum, Micah and Goldstein, Tom},
  journal={arXiv preprint arXiv:2306.04634},
  year={2023}
}

@inproceedings{wu2023llmdet,
  title={LLMDet: A Third Party Large Language Models Generated Text Detection Tool},
  author={Wu, Kangxi and Pang, Liang and Shen, Huawei and Cheng, Xueqi and Chua, Tat-Seng},
  booktitle={Findings of the Association for Computational Linguistics: EMNLP 2023},
  pages={2113--2133},
  year={2023}
}

@inproceedings{wu-etal-2025-wrote,
    title = "Who Wrote This? The Key to Zero-Shot {LLM}-Generated Text Detection Is {GECS}core",
    author = "Wu, Junchao  and
      Zhan, Runzhe  and
      Wong, Derek F.  and
      Yang, Shu  and
      Liu, Xuebo  and
      Chao, Lidia S.  and
      Zhang, Min",
    editor = "Rambow, Owen  and
      Wanner, Leo  and
      Apidianaki, Marianna  and
      Al-Khalifa, Hend  and
      Eugenio, Barbara Di  and
      Schockaert, Steven",
    booktitle = "Proceedings of the 31st International Conference on Computational Linguistics",
    month = jan,
    year = "2025",
    address = "Abu Dhabi, UAE",
    publisher = "Association for Computational Linguistics",
    url = "https://aclanthology.org/2025.coling-main.684/",
    pages = "10275--10292"
}

@inproceedings{nguyen2024simllm,
  title={SimLLM: Detecting sentences generated by large language models using similarity between the generation and its re-generation},
  author={Nguyen-Son, Hoang-Quoc and Dao, Minh-Son and Zettsu, Koji},
  booktitle={Proceedings of the 2024 Conference on Empirical Methods in Natural Language Processing},
  pages={22340--22352},
  year={2024}
}

@article{hu2023radar,
  title={Radar: Robust ai-text detection via adversarial learning},
  author={Hu, Xiaomeng and Chen, Pin-Yu and Ho, Tsung-Yi},
  journal={Advances in neural information processing systems},
  volume={36},
  pages={15077--15095},
  year={2023}
}

@article{guo-etal-2023-hc3,
    title = "How Close is ChatGPT to Human Experts? Comparison Corpus, Evaluation, and Detection",
    author = "Guo, Biyang  and
      Zhang, Xin  and
      Wang, Ziyuan  and
      Jiang, Minqi  and
      Nie, Jinran  and
      Ding, Yuxuan  and
      Yue, Jianwei  and
      Wu, Yupeng",
    journal={arXiv preprint arxiv:2301.07597},
    year = "2023",
}

@article{zhang2015character,
  title={Character-level convolutional networks for text classification},
  author={Zhang, Xiang and Zhao, Junbo and LeCun, Yann},
  journal={Advances in neural information processing systems},
  volume={28},
  year={2015}
}

@misc{qwen3technicalreport,
      title={Qwen3 Technical Report}, 
      author={Qwen Team},
      year={2025},
      eprint={2505.09388},
      archivePrefix={arXiv},
      primaryClass={cs.CL},
      url={https://arxiv.org/abs/2505.09388}, 
}

@article{qwen2.5-1m,
      title={Qwen2.5-1M Technical Report}, 
      author={An Yang and Bowen Yu and Chengyuan Li and Dayiheng Liu and Fei Huang and Haoyan Huang and Jiandong Jiang and Jianhong Tu and Jianwei Zhang and Jingren Zhou and Junyang Lin and Kai Dang and Kexin Yang and Le Yu and Mei Li and Minmin Sun and Qin Zhu and Rui Men and Tao He and Weijia Xu and Wenbiao Yin and Wenyuan Yu and Xiafei Qiu and Xingzhang Ren and Xinlong Yang and Yong Li and Zhiying Xu and Zipeng Zhang},
      journal={arXiv preprint arXiv:2501.15383},
      year={2025}
}

@inproceedings{wang-etal-2024-m4,
    title = "M4: Multi-generator, Multi-domain, and Multi-lingual Black-Box Machine-Generated Text Detection",
    author = "Wang, Yuxia  and
      Mansurov, Jonibek  and
      Ivanov, Petar  and
      Su, Jinyan  and
      Shelmanov, Artem  and
      Tsvigun, Akim  and
      Whitehouse, Chenxi  and
      Mohammed Afzal, Osama  and
      Mahmoud, Tarek  and
      Sasaki, Toru  and
      Arnold, Thomas  and
      Aji, Alham Fikri  and
      Habash, Nizar  and
      Gurevych, Iryna  and
      Nakov, Preslav",
    editor = "Graham, Yvette  and
      Purver, Matthew",
    booktitle = "Proceedings of the 18th Conference of the European Chapter of the Association for Computational Linguistics (Volume 1: Long Papers)",
    month = mar,
    year = "2024",
    address = "St. Julian{'}s, Malta",
    publisher = "Association for Computational Linguistics",
    url = "https://aclanthology.org/2024.eacl-long.83/",
    doi = "10.18653/v1/2024.eacl-long.83",
    pages = "1369--1407"
}

@article{liang2023gpt,
  title={GPT detectors are biased against non-native English writers},
  author={Liang, Weixin and Yuksekgonul, Mert and Mao, Yining and Wu, Eric and Zou, James},
  journal={Patterns},
  volume={4},
  number={7},
  year={2023},
  publisher={Elsevier}
}

@inproceedings{wahle-etal-2022-large,
    title = "How Large Language Models are Transforming Machine-Paraphrase Plagiarism",
    author = "Wahle, Jan Philip  and
      Ruas, Terry  and
      Kirstein, Frederic  and
      Gipp, Bela",
    editor = "Goldberg, Yoav  and
      Kozareva, Zornitsa  and
      Zhang, Yue",
    booktitle = "Proceedings of the 2022 Conference on Empirical Methods in Natural Language Processing",
    month = dec,
    year = "2022",
    address = "Abu Dhabi, United Arab Emirates",
    publisher = "Association for Computational Linguistics",
    url = "https://aclanthology.org/2022.emnlp-main.62/",
    doi = "10.18653/v1/2022.emnlp-main.62",
    pages = "952--963"
}

@article{pudasaini2024survey,
  title={Survey on AI-generated plagiarism detection: The impact of large language models on academic integrity},
  author={Pudasaini, Shushanta and Miralles-Pechu{\'a}n, Luis and Lillis, David and Llorens Salvador, Marisa},
  journal={Journal of Academic Ethics},
  pages={1--34},
  year={2024},
  publisher={Springer Netherlands}
}

@article{solaiman2019release,
  title={Release strategies and the social impacts of language models},
  author={Solaiman, Irene and Brundage, Miles and Clark, Jack and Askell, Amanda and Herbert-Voss, Ariel and Wu, Jeff and Radford, Alec and Krueger, Gretchen and Kim, Jong Wook and Kreps, Sarah and others},
  journal={arXiv preprint arXiv:1908.09203},
  year={2019}
}

@article{chen2025ipad,
  title={IPAD: Inverse Prompt for AI Detection--A Robust and Explainable LLM-Generated Text Detector},
  author={Chen, Zheng and Feng, Yushi and He, Changyang and Deng, Yue and Pu, Hongxi and Li, Bo},
  journal={arXiv preprint arXiv:2502.15902},
  year={2025}
}

@article{wu2024detectrl,
  title={Detectrl: Benchmarking llm-generated text detection in real-world scenarios},
  author={Wu, Junchao and Zhan, Runzhe and Wong, Derek and Yang, Shu and Yang, Xinyi and Yuan, Yulin and Chao, Lidia},
  journal={Advances in Neural Information Processing Systems},
  volume={37},
  pages={100369--100401},
  year={2024}
}

@misc{bao_glimpse_2025,
	title = {Glimpse: {Enabling} {White}-{Box} {Methods} to {Use} {Proprietary} {Models} for {Zero}-{Shot} {LLM}-{Generated} {Text} {Detection}},
	shorttitle = {Glimpse},
	url = {http://arxiv.org/abs/2412.11506},
	doi = {10.48550/arXiv.2412.11506},
	language = {en-US},
	urldate = {2025-09-23},
	publisher = {arXiv},
	author = {Bao, Guangsheng and Zhao, Yanbin and He, Juncai and Zhang, Yue},
	month = feb,
	year = {2025},
	note = {arXiv:2412.11506 [cs]},
}

@article{hermann2015teaching,
  title={Teaching machines to read and comprehend},
  author={Hermann, Karl Moritz and Kocisky, Tomas and Grefenstette, Edward and Espeholt, Lasse and Kay, Will and Suleyman, Mustafa and Blunsom, Phil},
  journal={Advances in neural information processing systems},
  volume={28},
  year={2015}
}

@article{zhou2025adadetectgpt,
  title={Adadetectgpt: Adaptive detection of llm-generated text with statistical guarantees},
  author={Zhou, Hongyi and Zhu, Jin and Su, Pingfan and Ye, Kai and Yang, Ying and Gavioli-Akilagun, Shakeel AOB and Shi, Chengchun},
  journal={arXiv preprint arXiv:2510.01268},
  year={2025}
}

@article{chen2024combating,
  title={Combating misinformation in the age of llms: Opportunities and challenges},
  author={Chen, Canyu and Shu, Kai},
  journal={AI Magazine},
  volume={45},
  number={3},
  pages={354--368},
  year={2024},
  publisher={Wiley Online Library}
}

@inproceedings{gressel2024discussion,
  title={Discussion paper: Exploiting llms for scam automation: A looming threat},
  author={Gressel, Gilad and Pankajakshan, Rahul and Mirsky, Yisroel},
  booktitle={Proceedings of the 3rd ACM Workshop on the Security Implications of Deepfakes and Cheapfakes},
  pages={20--24},
  year={2024}
}
\end{document}